%% file: main.tex
\documentclass{aifrontiers}
\usepackage{aifrontiers}
\usepackage{booktabs}
\usepackage{array}
\usepackage{colortbl}
\usepackage{caption}
\usepackage{subcaption}
\usepackage{multirow}
\usepackage{multicol}
\usepackage{float}
\usepackage{enumitem}
\usepackage{xspace}
\usepackage{soul}
\usepackage{listings}
\usepackage[export]{adjustbox}
\usepackage{algorithm}
\usepackage{algorithmic}

\input{preamble}

\begin{document}

\title{Learning Correct Behavior from Examples: Validating Sequential Execution in Autonomous Agents}
\shorttitle{Validating Sequential Execution in Autonomous Agents}
\author{
    Reshabh K Sharma$^{\spadesuit}$\thanks{Work done while at Microsoft.}, Gaurav Mittal$^{\heartsuit}$, Yu Hu$^{\heartsuit}$\\
    {\normalsize $^{\spadesuit}$University of Washington, Seattle, WA, USA \quad $^{\heartsuit}$Microsoft, Redmond, WA, USA}\\
    {\normalsize \texttt{reshabh@cs.washington.edu}, \texttt{Gaurav.Mittal@microsoft.com}, \texttt{yuhu@microsoft.com}}
}
\date{\today}

\renewcommand{\weblink}{}
\renewcommand{\foundrylink}{}
\renewcommand{\hflink}{}
\renewcommand{\ghlink}{}

\begin{abstract}
As autonomous agents become increasingly sophisticated, validating their sequential behavior presents a significant challenge. Traditional testing approaches require manual specification, exact sequence matching, or thousands of training examples. We present a novel algorithm that automatically learns correct behavior from just 2--10 passing execution traces and validates new executions against this learned model. Our approach combines dominator analysis from compiler theory with multimodal large language model-powered semantic understanding to identify essential states and handle non-deterministic behavior. The system constructs a generalized ground truth model using Prefix Tree Acceptors, merges traces through multi-tiered equivalence detection, and validates new executions via topological subsequence matching. In controlled experiments, our system achieved high accuracy in detecting product bugs and false successes using only 3 training traces. This approach provides explainable validation results with coverage metrics and works across diverse domains including UI testing, code generation, and robotic processes.
\end{abstract}

\maketitle

\input{introduction}

\input{example}

\input{design}
\input{algorithm}
\input{evaluation}
\input{discussion}

\input{limitation}

\input{related_work}

\input{conclusion}

\bibliographystyle{plainnat}
\bibliography{main}

\appendix
\input{prompts/sample}

\end{document}

%% file: preamble.tex
\usepackage[T1]{fontenc}    %
\usepackage[utf8]{inputenc} %

\newlength\savewidth

\newcolumntype{x}[1]{>{\centering\arraybackslash}p{#1pt}}
\newcolumntype{y}[1]{>{\raggedright\arraybackslash}p{#1pt}}

\definecolor{baselinecolor}{gray}{0.93}

\definecolor{bluebg}{RGB}{225, 235, 255}
\definecolor{bluetext}{RGB}{0, 0, 139}
\definecolor{orangebg}{RGB}{255, 228, 196}
\definecolor{orangetext}{RGB}{255, 69, 0}
\definecolor{purplebg}{RGB}{230, 230, 250}
\definecolor{purpletext}{RGB}{128, 0, 128}
\definecolor{redbg}{RGB}{255, 204, 204}
\definecolor{redtext}{RGB}{178, 34, 34}
\definecolor{greenbg}{RGB}{230, 255, 230}
\definecolor{greentext}{RGB}{0, 100, 0}
\definecolor{cyanbg}{RGB}{224, 255, 255}
\definecolor{cyantext}{RGB}{0, 139, 139}
\definecolor{yellowbg}{RGB}{255, 255, 204}
\definecolor{yellowtext}{RGB}{205, 133, 63}
\definecolor{violetbg}{RGB}{245, 230, 255}
\definecolor{violettext}{RGB}{148, 0, 211}
\definecolor{teal}{HTML}{1E88E5}
\definecolor{coral}{HTML}{D81B60}

\newcommand{\cellcolorpercent}[2]{
    \ifdim #1pt < #2pt
        \cellcolor{coral!15}#1\%
    \else
        \cellcolor{teal!15}#1\%
    \fi
    & 
    \ifdim #2pt < #1pt
        \cellcolor{coral!15}#2\%
    \else
        \cellcolor{teal!15}#2\%
    \fi
}

\newtcolorbox{quotefigurebox}{
  colframe=gray, colback=white, arc=5mm,
  boxrule=0.5pt, left=1pt, right=1pt, top=1pt, bottom=1pt,
  sharp corners,
}

%% file: introduction.tex
\section{Introduction}

As AI agents become more sophisticated, from computer use agents that interact with user interfaces~\cite{hu2024dawn} to coding agents that generate and refactor software~\cite{jimenez2024swebench}, an important question emerges. How do we know if an agent's behavior is correct? This challenge is particularly acute because autonomous systems rarely follow the exact same sequence of states and actions between executions. Loading screens may appear or disappear based on timing, alternative UI paths can accomplish the same goal, and different but equally valid code solutions can solve the same problem.

Traditional testing approaches exhibit limitations in handling these scenarios. Assertion-based testing requires manually writing assertions for every check, validates internal data but misses visual state issues, and cannot handle alternative execution paths~\cite{pezze2008software}. Record-and-replay tools are brittle, failing on minor rendering differences or timing variations~\cite{memon2016recordreplay}. Visual regression testing compares individual screenshots in isolation without understanding execution flow or semantic meaning~\cite{vitestVisualRegression,sparkbox2019visual}. Machine learning oracles require thousands of training examples and provide no explainability~\cite{fontes2021mloracles,braga2011mloracle}.

The core problem is \emph{non-determinism}~\cite{weyuker1982testing}. Autonomous systems exhibit variations in execution sequences due to timing differences, environmental factors, and legitimate alternative paths. Consider testing a computer use agent that needs to open VS Code and search for text in files. One execution might proceed by opening the Start Menu, typing ``VS Code,'' launching the application, displaying a loading screen, opening the search dialog, and finally showing results. Another execution might open the Start Menu, type ``VS Code,'' launch the application, and then proceed directly to the search dialog and results without any loading screen. The loading screen appears or disappears based on timing, but both executions are correct.

Traditional testing would either fail the second execution, being too brittle, or miss when truly essential steps are skipped, being too permissive. Consequently, there is a need for a system that can distinguish between acceptable variations and actual failures.

\subsection{Our Contribution}

We present a novel algorithm that automatically constructs a generalized ground truth model from a small number of passing execution traces and validates new executions by checking if they follow the essential structure of this model. Our key contributions are as follows.

\begin{itemize}
\item A three-phase algorithm combining Prefix Tree Acceptors~\cite{angluin1987learning} with dominator analysis~\cite{lengauer1979fast} to automatically identify essential versus optional states from 2--10 example traces
\item A multi-tiered state equivalence detection system that combines visual metrics with semantic LLM analysis to handle non-deterministic behavior
\item Topological subsequence matching that validates execution correctness while tolerating acceptable variations
\end{itemize}

Our approach is broadly applicable to any domain where sequential state transitions need validation, including UI testing, coding agents, robotic automation, and business process validation.

%% file: example.tex
\section{Motivating Example}

Consider a computer use agent that automates the task of opening VS Code and searching for specific text across files. We want to validate that the agent correctly completes this task across multiple executions.

\subsection{The Challenge}

Traditional testing approaches would struggle with this scenario. Exact sequence matching would fail because the loading screen may or may not appear depending on system performance. Manual assertions would require specifying every possible valid path, which is impractical. Pixel-perfect screenshot comparison would fail on minor variations in window decorations or font rendering.

\subsection{Our Approach}

We collect 3--5 passing execution traces where the agent successfully completes the task. Each trace captures sequential screenshots showing the UI state at each step along with actions taken between states such as clicks and keystrokes.

For example, traces might show different paths. In Trace 1, the agent opens the Start Menu, types ``VS Code,'' launches the application, sees a loading screen, reaches the main window, opens the search dialog, and displays results. In Trace 2, the agent follows the same sequence but proceeds directly from launch to the main window without any loading screen. Trace 3 resembles Trace 1, again showing the loading screen before the main window.

Our system automatically performs four key operations. First, it merges the traces into a unified graph structure with branches for alternative paths. Second, it identifies that the loading screen is optional because it appears in some traces but not others. Third, it extracts the dominator tree showing essential states: Start Menu, Launch, Main Window, Search Dialog, and Results. Fourth, it validates new executions by checking if they contain these essential states in order.

When a new test execution arrives, the system checks whether it follows the essential structure. If the agent skips directly from Launch to Search Dialog without ever showing the Main Window, this would be flagged as a failure because Main Window is an essential state. However, if the Loading Screen is missing, this is acceptable because it has been identified as optional.

This example illustrates the core challenge our work addresses. We aim to automatically learn which states are essential versus optional and validate new executions against this learned model without manual specification.

%% file: design.tex
\section{Design}
\label{sec:design}

\begin{figure}
    \centering
    \includegraphics[width=1\linewidth]{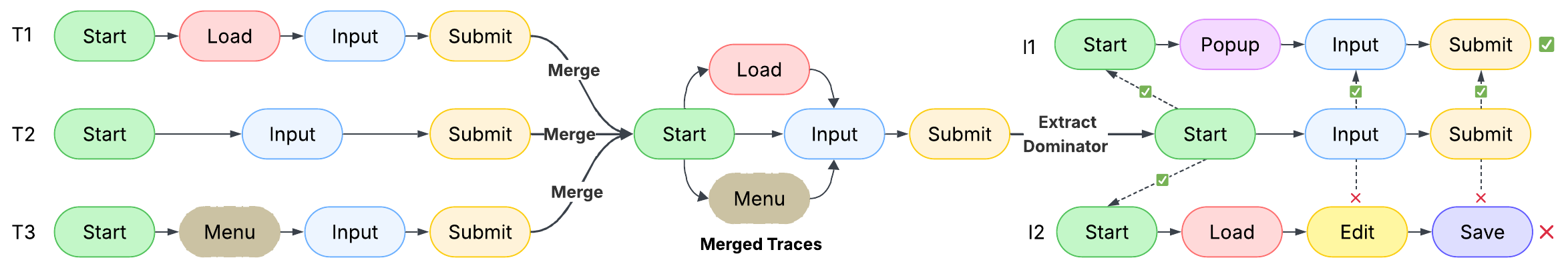}
    \caption{We illustrate the workings of our algorithm using three passing traces, T1, T2, and T3 (represented as PTAs), which are merged into a graph from which a dominator tree is extracted. The incoming traces I1 and I2 are then checked for a topological match with the dominator tree.}
    \label{fig:statemachineagent}
\end{figure}

Our algorithm consists of three phases as shown in Figure~\ref{fig:statemachineagent}. The first phase captures execution traces and constructs Prefix Tree Acceptors. The second phase merges traces and extracts dominator structure. The third phase validates new executions via topological subsequence matching.

\subsection{Phase 1: Capture Execution Traces}

We begin by collecting 2--10 execution traces from runs that are known to be correct. Each trace contains sequential state observations such as screenshots for UI agents, code snapshots for coding agents, or sensor readings for robotic systems. Additionally, each trace records actions or transitions including clicks, keystrokes, API calls, or other operations between states.

We convert each trace into a \emph{Prefix Tree Acceptor} (PTA)~\cite{angluin1987learning}, a directed graph where nodes represent observable states and edges represent actions. The state representation is flexible and domain-dependent, making the approach applicable across diverse domains. For this work, we focus on UI testing where states are captured as screenshots.

\subsection{Phase 2: Merge and Generalize}

Our primary contribution in this phase is merging multiple PTAs to handle non-determinism and identify essential versus optional states.

\subsubsection{Multi-Tiered State Equivalence Detection}

Determining when different screenshots represent the same logical UI state is challenging. We employ a three-tiered system.

\textbf{Tier 1} involves visual metrics. We compute fast perceptual comparisons including perceptual hash similarity~\cite{zauner2010phash} to measure visual similarity, structural similarity index (SSIM)~\cite{wang2004ssim} to determine structural alignment, and pixel change ratio to quantify differences. If all metrics clearly indicate equivalence by exceeding predefined thresholds, we merge the states.

\textbf{Tier 2} performs semantic analysis via LLM. When visual metrics are ambiguous, we invoke a multimodal large language model such as GPT-5.1 with a side-by-side comparison. We ask the model to analyze whether differences are semantically meaningful. Differences that are not meaningful include different window decorations, minor font rendering differences, and timestamp changes. Meaningful differences include different form validation errors, different data displayed, and different UI controls available. This semantic layer enables the system to understand functional equivalence beyond pixel-level comparison.

\textbf{Tier 3} handles smart merging with branches and convergence. As we merge PTAs, we build a graph structure that captures branches representing alternative execution paths such as those with or without loading screens. The structure also identifies convergence points where different paths rejoin, such as when all paths reach a ``save complete'' state.

\subsubsection{Dominator Extraction}

After merging, we compute which states ``dominate'' others~\cite{lengauer1979fast}. A state $d$ dominates state $s$ if every path from the initial state to $s$ must pass through $d$. For example, the initial state dominates everything because every execution starts there. A ``save'' action dominates the ``complete'' state because completion cannot occur without saving. Loading screens do not dominate anything because they are optional.

We extract a dominator subtree containing only essential states through a three-step process. We start with all terminal states representing successful endpoints. We then trace backward through immediate dominators to the initial state. Finally, we build a tree structure capturing the ``must-have'' execution flow. This automatically filters out optional variations while preserving the essential execution structure.

\subsection{Phase 3: Validate New Executions}

When a new test trace arrives, we extract its sequence of states and check whether it contains a topological subsequence that matches a path in the dominator tree.

Topological subsequence matching works as follows. If reference states are $A \rightarrow B \rightarrow C \rightarrow D$ and the test trace is $A \rightarrow X \rightarrow B \rightarrow Y \rightarrow Z \rightarrow C \rightarrow D$, this is a \textsc{match} because $A, B, C, D$ appear in correct order with $X, Y, Z$ as allowed extras.

We compute coverage metrics using the following formula.
\begin{equation}
\text{coverage} = \frac{\text{matched states}}{\text{total reference states}} \times 100\%
\end{equation}

The result is a \textbf{PASS} if coverage meets or exceeds the threshold (typically 100\%) and the terminal state matches. The result is a \textbf{FAIL} if essential states are missing or the wrong final state is reached. The system provides explainable results including coverage percentage, matched states, missing states, and a detailed explanation of the validation outcome.

%% file: algorithm.tex
\section{Algorithm}
\label{sec:algorithm}

This section presents the complete algorithm for learning correct behavior from execution traces and validating new executions. The algorithm consists of two main sub-algorithms: (1) extracting the dominator tree from passing traces (Algorithm~\ref{alg:extract}), and (2) evaluating test executions against the extracted dominator tree (Algorithm~\ref{alg:validate}).

\begin{algorithm}[t]
\caption{Extract Dominator Tree}
\label{alg:extract}
\begin{algorithmic}[1]
\REQUIRE $\mathcal{T} = \{T_1, T_2, \ldots, T_n\}$: set of $n$ passing execution traces
\ENSURE Dominator tree $D = (V_D, E_D)$ representing essential execution flow
\STATE \textbf{function} \textsc{ExtractDominatorTree}($\mathcal{T}$)
\STATE \COMMENT{Step 1: Construct Prefix Tree Acceptors for each trace}
\STATE $PTAs \leftarrow \emptyset$
\FORALL{trace $T_i \in \mathcal{T}$}
    \STATE $PTA_i \leftarrow$ \textsc{ConstructPTA}($T_i$)
    \STATE $PTAs \leftarrow PTAs \cup \{PTA_i\}$
\ENDFOR
\STATE \COMMENT{Step 2: Merge all PTAs into unified graph using multi-tiered state equivalence}
\STATE $G \leftarrow$ \textsc{MergePTAs}($PTAs$)
\STATE \COMMENT{Step 3: Compute dominator relationships using iterative dataflow analysis}
\STATE $dom \leftarrow$ \textsc{ComputeDominators}($G$)
\STATE \COMMENT{Step 4: Extract essential states by tracing back from terminal states}
\STATE $s_0 \leftarrow$ initial state of $G$
\STATE $T \leftarrow$ set of terminal states in $G$
\STATE $essential\_states \leftarrow \{s_0\}$
\STATE $V_D \leftarrow \{s_0\}$
\STATE $E_D \leftarrow \emptyset$
\FORALL{terminal state $t \in T$}
    \STATE $current \leftarrow t$
    \WHILE{$current \neq s_0$}
        \STATE $idom \leftarrow$ immediate dominator of $current$ in $dom$
        \IF{$idom \notin essential\_states$}
            \STATE $essential\_states \leftarrow essential\_states \cup \{idom\}$
            \STATE $V_D \leftarrow V_D \cup \{idom\}$
        \ENDIF
        \STATE $E_D \leftarrow E_D \cup \{(idom, current)\}$
        \STATE $current \leftarrow idom$
    \ENDWHILE
\ENDFOR
\STATE \COMMENT{Step 5: Build and return dominator tree}
\STATE $D \leftarrow (V_D, E_D)$
\RETURN $D$
\end{algorithmic}
\end{algorithm}

\begin{algorithm}[t]
\caption{Validate Execution}
\label{alg:validate}
\begin{algorithmic}[1]
\REQUIRE Test trace $T_{test}$, dominator tree $D$, coverage threshold $\theta$
\ENSURE Validation result: PASS/FAIL with coverage metrics and explanation
\STATE \textbf{function} \textsc{ValidateExecution}($T_{test}$, $D$, $\theta$)
\STATE \COMMENT{Extract states from test trace}
\STATE $S_{test} \leftarrow \langle s_1^{test}, s_2^{test}, \ldots, s_m^{test} \rangle$ from $T_{test}$
\STATE \COMMENT{Get reference states from dominator tree in topological order}
\STATE $S_{ref} \leftarrow$ \textsc{TopologicalOrder}($D$)
\STATE \COMMENT{Perform topological subsequence matching using state equivalence}
\STATE $(matched, missing) \leftarrow$ \textsc{TopologicalSubsequenceMatch}($S_{test}$, $S_{ref}$)
\STATE \COMMENT{Compute coverage metrics}
\STATE $coverage \leftarrow |matched| / |S_{ref}| \times 100\%$
\STATE $terminal_{ref} \leftarrow$ terminal state in $D$
\STATE $terminal_{test} \leftarrow$ last state in $S_{test}$
\STATE $terminal\_match \leftarrow$ \textsc{StatesEquivalent}($terminal_{test}$, $terminal_{ref}$)
\STATE \COMMENT{Make validation decision}
\IF{$coverage \geq \theta$ \textbf{and} $terminal\_match$}
    \RETURN (\textsc{PASS}, $coverage$, $matched$, $\emptyset$)
\ELSE
    \STATE $explanation \leftarrow$ ``Missing essential states: '' + $missing$ + ``. Coverage: '' + $coverage$
    \RETURN (\textsc{FAIL}, $coverage$, $matched$, $missing$, $explanation$)
\ENDIF
\end{algorithmic}
\end{algorithm}

\subsection{Complexity Analysis}

The time complexity of the algorithm is dominated by the merging phase. For $n$ traces with average length $k$, PTA construction runs in $O(n \cdot k)$ time, which is linear in trace length. State equivalence checking requires $O(n^2 \cdot k^2)$ time in the worst case, but typically performs much better with early termination in Tier 1. Dominator extraction runs in $O(|V_G| + |E_G|)$ time using standard dominator tree algorithms~\cite{lengauer1979fast}. Validation requires $O(m \cdot n)$ time where $m$ is the test trace length and $n$ is the number of reference states.

The space complexity is $O(n \cdot k)$ for storing the merged graph and dominator tree.

%% file: evaluation.tex
\section{Case Study: VS Code Extension Bug Detection}
\label{sec:evaluation}

While a large-scale empirical study is reserved for future work, we designed a targeted, controlled case study to validate the core mechanics of our algorithm. This preliminary evaluation uses a controlled synthetic benchmark comprising 28 agent executions that mimic real-world testing conditions, allowing precise measurement of detection accuracy across failure categories.

\subsection{Experimental Setup}

We created a synthetic bug test using a custom VS Code extension. The passing configuration consisted of one VM with the custom extension installed, producing correct execution traces. The failing configuration used another VM without the extension installed, simulating a product bug scenario. For model training, we used 3 passing traces to build the dominator tree that serves as the trace model.

We compare our approach against a baseline where the Computer Use Agent (CUA) self-reports its own success or failure. This baseline represents the common practice of relying on an agent's internal assessment rather than independent structural validation.

\subsection{Research Questions}

We designed our evaluation to answer the following research questions.

\textbf{RQ1: Can the system accurately detect different types of failures?} We evaluate the system's ability to distinguish between passing traces, false successes, agent issues, product bugs, and missed bugs.

\textbf{RQ2: How does structural validation compare to agent self-assessment?} We compare the accuracy, precision, recall, and F1-score of our dominator tree approach against the CUA's internal success reporting.

\textbf{RQ3: Can the system identify ``not a bug'' scenarios?} Beyond detecting failures, we investigate whether the system can correctly identify when a test failure is due to agent execution error rather than an actual product regression.

\subsection{Results}

Using the dominator tree built from 3 passing traces, we evaluated the system on the remaining 25 traces comprising 14 failing traces (3 agent issues and 11 product bugs) and 11 passing traces. Additionally, we assessed the system's ability to catch CUA misclassifications: cases where the CUA incorrectly reported success on a failing trace (false success) or incorrectly reported failure on a passing trace (missed bug). The system achieved 100\% detection accuracy across all categories: false successes (1/1), agent issues (3/3), product bugs (11/11), and missed bugs (1/1). Table~\ref{tab:comparison} compares our approach against CUA self-assessment, and Table~\ref{tab:classification} shows classification accuracy when distinguishing failure root causes.

\begin{table}[h]
\centering
\begin{minipage}[t]{0.48\linewidth}
\centering
\begin{tabular}{p{0.30\linewidth} p{0.25\linewidth}p{0.35\linewidth}}
\toprule
\textbf{Metric} & \textbf{CUA} & \textbf{Ours} \\
\midrule
Accuracy & 82.2\% & 100\% (+17.8pp) \\
Precision & 83.3\% & 100\% (+16.7pp) \\
Recall & 60.0\% & 100\% (+40.0pp) \\
F1-Score & 69.8\% & 100\% (+30.2pp) \\
\bottomrule
\end{tabular}
\caption{CUA self-assessment vs.\ dominator tree validation}
\label{tab:comparison}
\end{minipage}
\hfill
\begin{minipage}[t]{0.48\linewidth}
\centering
\begin{tabular}{l c}
\toprule
\textbf{Trace Type} & \textbf{Accuracy} \\
\midrule
Agent Issue & 33.3\% (1/3) \\
Product Bug & 72.7\% (8/11) \\
\bottomrule
\end{tabular}
\caption{Classification accuracy for failure root causes}
\label{tab:classification}
\end{minipage}
\end{table}

\textbf{RQ1 Results:} The system achieved 100\% accuracy on our synthetic product bug subset (11/11) and 100\% accuracy across all other categories. This demonstrates that the approach can effectively learn correct behavior from just 3 passing traces and use this model to validate new executions.

\textbf{RQ2 Results:} Our structural validation approach substantially outperforms CUA self-assessment across all metrics (Table~\ref{tab:comparison}). The CUA frequently misreported failures as successes, often due to timing out or misinterpreting its own state, achieving only 82.2\% accuracy and 60.0\% recall. In contrast, the dominator tree achieved perfect differentiation by focusing on whether essential milestones were actually reached rather than relying on the agent's internal assessment.

\textbf{RQ3 Results:} The most significant impact for developers is in reducing false alarms. When a test fails, high-signal feedback is needed to determine if the product code is broken or if the agent simply stumbled due to environmental noise. The CUA's internal self-assessment was completely unable to identify ``not a bug'' scenarios (0\% F1-score), demonstrating that agents cannot yet reliably assess their own performance in non-deterministic environments. By using state and action equivalence within the dominator model, our approach achieved a 52.2\% F1-score in correctly identifying when a failure was an agent execution error rather than a product regression. The system correctly identified 1 of 3 agent issues and 8 of 11 product bugs (Table~\ref{tab:classification}). While there is room for improvement, this capability significantly reduces manual review time wasted on flaky test results and false positives in CI pipelines.

\subsection{Threats to Validity}

Our evaluation uses a synthetic bug scenario with a controlled VS Code extension, which allows precise measurement but may not capture the complexity of real-world failure patterns; additionally, the small sample sizes (e.g.,\ 1 false success, 1 missed bug) limit statistical confidence for some categories. Results from UI testing with computer use agents may not generalize to all domains, and the reliance on visual state representation assumes screenshot-based observation, meaning domains with non-visual state such as backend services would require alternative representations. Finally, our accuracy metric treats all failure types equally, though in practice some failures such as missed bugs may be more costly than others, and the distinction between agent issues and product bugs depends on ground truth labeling that may be ambiguous in edge cases.

%% file: discussion.tex
\section{Discussion}

The strong performance on product bug detection makes this approach particularly valuable for regression testing~\cite{yoo2012regression} and continuous integration pipelines~\cite{hilton2016ci}, where quickly identifying genuine failures is important. The ability to learn from just 3 passing traces significantly reduces setup cost compared to traditional machine learning approaches that require hundreds or thousands of examples. The system's detection of all false successes and missed bugs in our test set demonstrates its ability to catch subtle failures that might pass traditional assertion-based tests---a capability particularly relevant for validating autonomous agents where the execution path matters as much as the final outcome.

Our approach provides several key advantages over existing validation techniques. Traditional assertion-based testing requires developers to manually write assertions for every check and cannot handle alternative execution paths; our approach eliminates this manual effort by automatically learning from examples and naturally accommodates multiple valid paths through merged graph structures. Machine learning-based test oracles require thousands of training examples and operate as black boxes, while our approach achieves effective validation with just 2--10 traces and produces an interpretable dominator tree with explainable results. Record-and-replay tools fail on minor rendering differences and require exact matches; our multi-tiered equivalence detection combines visual metrics with LLM semantic analysis to tolerate acceptable variations while generalizing from multiple traces. Finally, symbolic execution and formal verification methods require source code access and suffer from path explosion; our black-box approach learns from observed executions and uses dominator extraction to reduce state space complexity.

While our evaluation focuses on UI testing with computer use agents, the approach applies broadly to other domains. For coding agents~\cite{jimenez2024swebench}, the system can learn essential development patterns (e.g.,\ creating tests, implementing features, verifying tests pass) and validate that AI-generated solutions follow these checkpoints even when implementation details vary. For robotic process automation~\cite{vanderaalst2018rpa}, the dominator tree captures mandatory compliance checkpoints and approval gates that must occur regardless of conditional branching. For reinforcement learning from demonstrations~\cite{argall2009demonstrations}, the approach provides quality assurance for training data by filtering demonstration traces that skip essential waypoints.

The integration of multimodal LLM-based semantic analysis addresses ambiguous cases where visual metrics alone are insufficient. This semantic layer distinguishes functionally irrelevant differences (e.g.,\ window decorations) from meaningful ones (e.g.,\ error messages). As LLM capabilities continue to improve, equivalence detection accuracy should increase accordingly.

%% file: limitation.tex
\section{Limitations and Future Work}

The system requires passing traces and cannot learn from failures alone, though obtaining a few successful runs is typically straightforward in practice. The current implementation relies on visual state representation, making it ideal for UI testing but less effective for pure backend services; non-visual domains would require alternative state representations such as API responses or database states. Semantic equivalence checking introduces an LLM dependency with associated API costs, though visual metrics alone provide a fallback. The current implementation also does not model timing information or temporal constraints.

Several directions could extend this work. Temporal constraint modeling would enable validation of performance-critical behavior by learning acceptable time bounds for operations. Learning from negative examples could improve discrimination by identifying divergence points where failures occur. Hierarchical state abstraction could cluster low-level states into high-level concepts (e.g.,\ abstracting multiple launch screenshots into a single ``application launch'' state). Multi-modal state representation combining screenshots with DOM structure, accessibility trees, or network traffic could improve equivalence detection. Online learning that updates the dominator tree from new validated traces would enable continuous model refinement.

%% file: related_work.tex
\section{Related Work}

This research bridges classical state-machine inference and modern AI validation. We contextualize our dominator-based approach against three relevant domains: software testing, automata learning, and compiler theory. While these fields provide foundational techniques for structural analysis and behavioral modeling, we highlight their limitations in handling the non-determinism inherent in autonomous agent execution.

\subsection{Software Testing and Validation}

Traditional software testing approaches include unit testing with assertions, integration testing, and end-to-end testing~\cite{pezze2008software}. These methods require manual specification of expected behavior and struggle with non-deterministic systems~\cite{weyuker1982testing}. Our work complements these approaches by providing automatic learning of expected behavior from examples.

Record-and-replay testing tools capture user interactions and play them back to detect regressions~\cite{memon2016recordreplay}. However, these tools are brittle and fail on minor variations. Our approach generalizes from multiple traces and uses semantic equivalence to tolerate acceptable variations.

Visual regression testing tools compare screenshots to detect UI changes~\cite{sparkbox2019visual,vitestVisualRegression}. Unlike these tools that compare states in isolation, our approach validates execution flow and understands the sequential dependencies between states.

\subsection{Machine Learning for Testing}

Machine learning-based test oracles (e.g.,\ neural network classifiers~\cite{fontes2021mloracles,braga2011mloracle,malaiya2004neuraloracle}) learn pass/fail classification from large datasets but require extensive training data and provide no explainability. Our work achieves effective validation with minimal examples and produces interpretable structural models. Metamorphic testing~\cite{segura2016surveyMT,chen2018mtChallenges} generates test cases using system properties without requiring a test oracle, but demands domain-specific metamorphic relations; our approach learns automatically from examples.

\subsection{Formal Methods and Model-Based Testing}

Symbolic execution and model checking enumerate execution paths and verify properties~\cite{cadar2013symbolic,clarke1999model}. These approaches suffer from path explosion and require source code access. Our black-box approach learns from observed behavior and uses dominator analysis to manage state space complexity.

Model-based testing requires manual construction of state machine models~\cite{utting2007practicalMBT,utting2012taxonomyMBT}. Our work automates model construction from execution traces, eliminating the need for manual modeling expertise.

\subsection{Automata Learning}

Automata learning techniques for inferring state machine models fall into two categories, each with distinct limitations that our dominator-based approach addresses.

\textbf{Active learning approaches} such as L*~\cite{angluin1987learning} and EFSM inference~\cite{walkinshaw2016efsmInference} require extensive system interaction through membership and equivalence queries. This assumption breaks down for autonomous agents where querying is expensive, slow, or impractical---for instance, a computer-use agent interacting with production UIs cannot be queried thousands of times to infer a model. Our approach instead learns passively from 2--10 observed traces, requiring no system interaction.

\textbf{Passive learning approaches} construct models from traces without interaction but face different challenges. SAT-based inference~\cite{avellaneda2018fsmLong} builds FSMs iteratively from trace subsets, while invariant-enhanced mining~\cite{krka2014mining} augments traces with program invariants. Both produce complete state machines that grow with trace complexity and do not distinguish essential states from optional variations. In contrast, our dominator extraction automatically identifies which states are structurally necessary---states that every successful execution must visit---versus optional states like loading screens that may or may not appear. This distinction, adapted from compiler control-flow analysis~\cite{dietl2007runtime}, enables validation that tolerates acceptable non-determinism while enforcing essential behavior.

\subsection{Compiler Theory and Program Analysis}

Dominator analysis is a well-established technique in compiler optimization and program analysis~\cite{lengauer1979fast,cooper2001simple}. We adapt this technique to the domain of execution trace validation, using it to identify essential states in behavioral sequences rather than control flow in programs.

\subsection{Autonomous Agents and AI Testing}

As autonomous agents become more prevalent, validating their behavior becomes increasingly important. Recent work on testing AI systems focuses on adversarial testing and robustness evaluation~\cite{goodfellow2015explaining,carlini2017towards}. Our work addresses the complementary problem of validating correct sequential behavior in non-adversarial settings.

Several recent approaches address agent validation through specification-based checking. ContextCov~\cite{sharma2026contextcov} synthesizes executable checks from natural language agent instructions (e.g.,\ AGENTS.md files) to enforce coding conventions and architectural constraints; however, this approach requires explicit natural language specifications, which do not exist for the visual sequential behaviors we target. AgentPex~\cite{sharma2026willful} extracts behavioral rules from agent prompts and evaluates chat agent traces for compliance, demonstrating that agents frequently violate their own instructions---a finding that strengthens the case for automated validation approaches like ours. AgentRx~\cite{barke2026agentrx} contributes a cross-domain failure taxonomy and diagnostic framework for tool-using agents, synthesizing constraints from tool schemas and policies to localize the first unrecoverable failure; however, AgentRx relies on manually annotated failure benchmarks and domain-specific tool schemas, whereas our approach learns validation models automatically from a small number of passing traces without manual annotation.

%% file: conclusion.tex
\section{Conclusion}

The proliferation of autonomous agents~\cite{wang2024survey} demands validation techniques that can accommodate non-deterministic execution without sacrificing rigorous correctness. This paper demonstrates that reliable behavioral validation does not require massive training corpora, brittle exact-match specifications, or manual assertions.

We introduced a novel automated validation framework that extracts generalized ground-truth models from just 2--10 passing execution traces. By integrating multimodal LLM semantic equivalence with dominator analysis~\cite{lengauer1979fast,cooper2001simple}, our system successfully isolates essential execution states from acceptable, non-deterministic variations. Our preliminary evaluation demonstrates the efficacy of this approach: using a model built from only three passing traces, the system achieved 100\% accuracy in detecting product bugs and successfully identified false successes while tolerating valid UI variations.

Furthermore, the reliance on dominator trees ensures that our validation results remain fully explainable, generating precise coverage metrics rather than black-box classifications. As autonomous agents are increasingly deployed in complex production environments, our approach provides a practical, low-shot, and interpretable mechanism for ensuring their reliability.

%% file: prompts/sample.tex
\section{Appendix: Semantic Equivalence Prompt}

\subsection{LLM Prompt for State Equivalence Detection}
\label{prompt:equivalence}
\begin{lstlisting}
Compare these two UI screenshots side-by-side.
Are the differences semantically meaningful?

Examples of NOT meaningful:
- Different window decorations
- Minor font rendering differences
- Timestamp changes

Examples of MEANINGFUL:
- Different form validation errors
- Different data displayed
- Different UI controls available

Please analyze the images and respond with:
1. Whether the differences are semantically meaningful (Yes/No)
2. A brief explanation of the key differences
3. Your confidence level in this assessment

Response format:
{
  "equivalent": true/false,
  "explanation": "...",
  "confidence": "high/medium/low"
}
\end{lstlisting}

%% file: main.bib
@book{pezze2008software,
  title={Software testing and analysis: process, principles, and techniques},
  author={Pezz{\`e}, Mauro and Young, Michal},
  year={2008},
  publisher={John Wiley \& Sons}
}

@article{weyuker1982testing,
  title={On testing non-testable programs},
  author={Weyuker, Elaine J},
  journal={The Computer Journal},
  volume={25},
  number={4},
  pages={465--470},
  year={1982},
  publisher={The British Computer Society}
}

@inproceedings{memon2016recordreplay,
  title={Regression testing of web applications using record/replay tools},
  author={Hammoudi, Mouna},
  booktitle={Proceedings of the 2016 24th ACM SIGSOFT International Symposium on Foundations of Software Engineering},
  pages={1079--1081},
  year={2016}
}

@misc{vitestVisualRegression,
  title        = {Visual Regression Testing},
  howpublished = {\url{https://main.vitest.dev/guide/browser/visual-regression-testing}},
  note         = {Accessed: 2026-01-21}
}

@misc{sparkbox2019visual,
  title        = {Visual Regression Testing in Design Systems},
  howpublished = {\url{https://sparkbox.com/foundry/design_system_visual_regression_testing}},
  note         = {Accessed: 2026-01-21}
}

@inproceedings{fontes2021mloracles,
  title={Using machine learning to generate test oracles: A systematic literature review},
  author={Fontes, Afonso and Gay, Gregory},
  booktitle={Proceedings of the 1st International Workshop on Test Oracles},
  pages={1--10},
  year={2021}
}

@inproceedings{braga2011mloracle,
  title={A machine learning approach to generate test oracles},
  author={Braga, Rony{\'e}rison and Neto, Pedro Santos and Rab{\^e}lo, Ricardo and Santiago, Jos{\'e} and Souza, Matheus},
  booktitle={Proceedings of the XXXII Brazilian Symposium on Software Engineering},
  pages={142--151},
  year={2018}
}

@article{malaiya2004neuraloracle,
  title={A neural net based approach to test oracle},
  author={Aggarwal, KK and Singh, Yogesh and Kaur, Arvinder and Sangwan, OP},
  journal={ACM SIGSOFT Software Engineering Notes},
  volume={29},
  number={3},
  pages={1--6},
  year={2004},
  publisher={ACM New York, NY, USA}
}

@article{segura2016surveyMT,
  title={A survey on metamorphic testing},
  author={Segura, Sergio and Fraser, Gordon and Sanchez, Ana B and Ruiz-Cort{\'e}s, Antonio},
  journal={IEEE Transactions on software engineering},
  volume={42},
  number={9},
  pages={805--824},
  year={2016},
  publisher={IEEE}
}

@article{chen2018mtChallenges,
  title={Metamorphic testing: A review of challenges and opportunities},
  author={Chen, Tsong Yueh and Kuo, Fei-Ching and Liu, Huai and Poon, Pak-Lok and Towey, Dave and Tse, TH and Zhou, Zhi Quan},
  journal={ACM Computing Surveys (CSUR)},
  volume={51},
  number={1},
  pages={1--27},
  year={2018},
  publisher={ACM New York, NY, USA}
}

@article{cadar2013symbolic,
  title={Symbolic execution for software testing: three decades later},
  author={Cadar, Cristian and Sen, Koushik},
  journal={Communications of the ACM},
  volume={56},
  number={2},
  pages={82--90},
  year={2013},
  publisher={ACM New York, NY, USA}
}

@book{clarke1999model,
  title={Model checking},
  author={Clarke, Edmund M},
  booktitle={International conference on foundations of software technology and theoretical computer science},
  pages={54--56},
  year={1997},
  organization={Springer}
}

@book{utting2007practicalMBT,
  title={Practical model-based testing: a tools approach},
  author={Utting, Mark and Legeard, Bruno},
  year={2010},
  publisher={Elsevier}
}

@article{utting2012taxonomyMBT,
  title={A taxonomy of model-based testing approaches},
  author={Utting, Mark and Pretschner, Alexander and Legeard, Bruno},
  journal={Software testing, verification and reliability},
  volume={22},
  number={5},
  pages={297--312},
  year={2012},
  publisher={Wiley Online Library}
}

@article{angluin1987learning,
  title={Learning regular sets from queries and counterexamples},
  author={Angluin, Dana},
  journal={Information and computation},
  volume={75},
  number={2},
  pages={87--106},
  year={1987},
  publisher={Elsevier}
}

@article{walkinshaw2016efsmInference,
  title={Inferring extended finite state machine models from software executions},
  author={Walkinshaw, Neil and Taylor, Ramsay and Derrick, John},
  journal={Empirical software engineering},
  volume={21},
  number={3},
  pages={811--853},
  year={2016},
  publisher={Springer}
}

@inproceedings{dietl2007runtime,
  author    = {Dietl, Werner and M{\"u}ller, Peter},
  title={Runtime universe type inference},
  author={Dietl, Werner and M{\"u}ller, Peter},
  booktitle={International Workshop on Aliasing, Confinement and Ownership in object-oriented programming (IWACO)},
  pages={72--80},
  year={2007}
}

@inproceedings{krka2014mining,
  title={Automatic mining of specifications from invocation traces and method invariants},
  author={Krka, Ivo and Brun, Yuriy and Medvidovic, Nenad},
  booktitle={Proceedings of the 22nd ACM SIGSOFT International Symposium on Foundations of Software Engineering},
  pages={178--189},
  year={2014}
}

@inproceedings{avellaneda2018fsmLong,
  title={FSM inference from long traces},
  author={Avellaneda, Florent and Petrenko, Alexandre},
  booktitle={International Symposium on Formal Methods},
  pages={93--109},
  year={2018},
  organization={Springer}
}

@article{lengauer1979fast,
  title={A fast algorithm for finding dominators in a flowgraph},
  author={Lengauer, Thomas and Tarjan, Robert Endre},
  journal={ACM Transactions on Programming Languages and Systems (TOPLAS)},
  volume={1},
  number={1},
  pages={121--141},
  year={1979},
  publisher={ACM New York, NY, USA}
}

@article{cooper2001simple,
  title={A simple, fast dominance algorithm},
  author={Cooper, Keith D and Harvey, Timothy J and Kennedy, Ken},
  journal={Software Practice \& Experience},
  volume={4},
  number={1-10},
  pages={1--8},
  year={2001}
}

@article{sharma2026contextcov,
  title={ContextCov: Deriving and Enforcing Executable Constraints from Agent Instruction Files},
  author={Sharma, Reshabh K},
  journal={arXiv preprint arXiv:2603.00822},
  year={2026}
}

@article{sharma2026willful,
  title={Willful Disobedience: Automatically Detecting Failures in Agentic Traces},
  author={Sharma, Reshabh K and Barke, Shraddha and Zorn, Benjamin},
  journal={arXiv preprint arXiv:2603.23806},
  year={2026}
}

@article{barke2026agentrx,
  title={AgentRx: Diagnosing AI Agent Failures from Execution Trajectories},
  author={Barke, Shraddha and Goyal, Arnav and Khare, Alind and Singh, Avaljot and Nath, Suman and Bansal, Chetan},
  journal={arXiv preprint arXiv:2602.02475},
  year={2026}
}

@misc{goodfellow2015explaining,
      title={Explaining and Harnessing Adversarial Examples}, 
      author={Ian J. Goodfellow and Jonathon Shlens and Christian Szegedy},
      year={2015},
      eprint={1412.6572},
      archivePrefix={arXiv},
      primaryClass={stat.ML},
      url={https://arxiv.org/abs/1412.6572}, 
}

@misc{carlini2017towards,
      title={Towards Evaluating the Robustness of Neural Networks}, 
      author={Nicholas Carlini and David Wagner},
      year={2017},
      eprint={1608.04644},
      archivePrefix={arXiv},
      primaryClass={cs.CR},
      url={https://arxiv.org/abs/1608.04644}, 
}

@article{hu2024dawn,
  title={The dawn of gui agent: A preliminary case study with claude 3.5 computer use},
  author={Hu, Siyuan and Ouyang, Mingyu and Gao, Difei and Shou, Mike Zheng},
  journal={arXiv preprint arXiv:2411.10323},
  year={2024}
}

@inproceedings{jimenez2024swebench,
      title={SWE-bench: Can Language Models Resolve Real-World GitHub Issues?}, 
      author={Carlos E. Jimenez and John Yang and Alexander Wettig and Shunyu Yao and Kexin Pei and Ofir Press and Karthik Narasimhan},
      year={2024},
      eprint={2310.06770},
      archivePrefix={arXiv},
      primaryClass={cs.CL},
      url={https://arxiv.org/abs/2310.06770}, 
}

@article{wang2004ssim,
  title={Image quality assessment: from error visibility to structural similarity},
  author={Wang, Zhou and Bovik, Alan C and Sheikh, Hamid R and Simoncelli, Eero P},
  journal={IEEE Transactions on Image Processing},
  volume={13},
  number={4},
  pages={600--612},
  year={2004},
  publisher={IEEE}
}

@inproceedings{zauner2010phash,
  title={Implementation and Benchmarking of Perceptual Image Hash Functions},
  author={Christoph Zauner},
  year={2010},
  url={https://api.semanticscholar.org/CorpusID:17075066}
}

@article{wang2024survey,
  title={A survey on large language model based autonomous agents},
  author={Wang, Lei and Ma, Chen and Feng, Xueyang and Zhang, Zeyu and Yang, Hao and Zhang, Jingsen and Chen, Zhiyuan and Tang, Jiakai and Chen, Xu and Lin, Yankai and others},
  journal={Frontiers of Computer Science},
  volume={18},
  number={6},
  pages={186345},
  year={2024},
  publisher={Springer}
}

@article{yoo2012regression,
  title={Regression testing minimization, selection and prioritization: a survey},
  author={Yoo, Shin and Harman, Mark},
  journal={Software testing, verification and reliability},
  volume={22},
  number={2},
  pages={67--120},
  year={2012},
  publisher={Wiley Online Library}
}

@inproceedings{hilton2016ci,
  title={Usage, costs, and benefits of continuous integration in open-source projects},
  author={Hilton, Michael and Tunnell, Timothy and Huang, Kai and Marinov, Darko and Dig, Danny},
  booktitle={Proceedings of the 31st IEEE/ACM International Conference on Automated Software Engineering},
  pages={426--437},
  year={2016}
}

@article{vanderaalst2018rpa,
  title={Robotic process automation},
  author={Van der Aalst, Wil MP and Bichler, Martin and Heinzl, Armin},
  journal={Business \& Information Systems Engineering},
  volume={60},
  number={4},
  pages={269--272},
  year={2018},
  publisher={Springer}
}

@article{argall2009demonstrations,
  title={A survey of robot learning from demonstration},
  author={Argall, Brenna D and Chernova, Sonia and Veloso, Manuela and Browning, Brett},
  journal={Robotics and autonomous systems},
  volume={57},
  number={5},
  pages={469--483},
  year={2009},
  publisher={Elsevier}
}
